%% file: 0536.tex
\documentclass[runningheads]{llncs}
\usepackage{graphicx}

\usepackage{tikz}
\usepackage{comment} 
\usepackage{amsmath,amssymb} 
\usepackage{color}
\usepackage{float}
\usepackage{amsfonts}
\usepackage{subfigure}
\usepackage{booktabs}

\newcommand{\inc}[1]{\textcolor{red}{#1}}
\newcommand{\dec}[1]{\textcolor{blue}{#1}}
\newcommand{\gre}[1]{\textcolor[rgb]{0,0.6,0.3}{#1}}

\begin{document}
\pagestyle{headings}
\mainmatter
\def\ECCVSubNumber{536}  

\title{Cheaper Pre-training Lunch: An Efficient Paradigm for Object Detection} 

\titlerunning{Cheaper Pre-training Lunch}
%
\author{Dongzhan~Zhou\inst{1}\thanks{Co-first author.} \and
Xinchi~Zhou\inst{1}$^\star$ \and
Hongwen~Zhang\inst{2} \and
Shuai~Yi\inst{3} \and
Wanli~Ouyang\inst{1}
}
\authorrunning{Zhou et al.}
%
\institute{The University of Sydney, SenseTime Computer Vision Research Group, Australia \email{\{d.zhou,xinchi.zhou1,wanli.ouyang\}@sydney.edu.au } \and
Institute of Automation, Chinese Academy of Sciences \& University of Chinese Academy of Sciences\\
\email{hongwen.zhang@cripac.ia.ac.cn} \and
Sensetime Research\\
\email{yishuai@sensetime.com}}
\maketitle

\begin{abstract}
\input{contents/abstract.tex}
\keywords{Pre-training, Object Detection, Acceleration, Deep Neural Networks, Deep Learning}
\end{abstract}

\section{Introduction}
\input{contents/introduction.tex}

\section{Related Work}
\input{contents/related_works.tex}

\section{Methodology}
\input{contents/dataset.tex}

\input{contents/jigsaw.tex}
\input{contents/losses.tex}

\section{Experiments}
\input{contents/experiments.tex}

\section{Discussion}
\input{contents/discussion.tex}

\section{Conclusions}
\input{contents/conclusion.tex}
\\
\noindent\textbf{Acknowledgement} This work was supported by SenseTime, the Australian Research Council Grant  DP200103223, and Australian Medical Research Future Fund MRFAI000085.

\clearpage
\section*{Appendix}
\appendix
\input{contents/supp.tex}
\clearpage
%
%
\bibliographystyle{splncs04}
\bibliography{egbib}
\end{document}

%% file: contents/abstract.tex
In this paper, we propose a general and efficient pre-training paradigm, Montage pre-training, for object detection. Montage pre-training needs only the target detection dataset while taking only $1/4$ computational resources compared to the widely adopted ImageNet pre-training. To build such an efficient paradigm, we reduce the potential redundancy by carefully extracting useful samples from the original images, assembling samples in a Montage manner as input, and using an ERF-adaptive dense classification strategy for model pre-training. These designs include not only a new input pattern to improve the spatial utilization but also a novel learning objective to expand the effective receptive field of the pre-trained model. The efficiency and effectiveness of Montage pre-training are validated by extensive experiments on the MS-COCO dataset, where the results indicate that the models using Montage pre-training are able to achieve on-par or even better detection performances compared with the ImageNet pre-training.

%% file: contents/introduction.tex
Pre-training on the classification dataset (\textit{e.g.}~ImageNet~\cite{krizhevsky2012imagenet}) is a common practice to achieve better network initialization for object detection. Under this paradigm, deep networks benefit from useful feature representations learned from large-scale data, which promotes the convergence of models during fine-tuning stage. Despite the benefits, the burdens caused by extra data should not be neglected. 

Previous works~\cite{shen2017dsod,he2019rethinking,zhu2019scratchdet} have proposed alternative solutions to directly train detection models from scratch with random initialization. However, there is always no free lunch. Training from scratch suffers from slower convergence, namely, additional training iterations are needed to obtain competitive models. \emph{Can we incorporate the merit of fast convergence via pre-training without paying for the extra data or expensive training cost?}

The answer is \textbf{Yes}. We find the cheaper lunch for pre-training. In this work, we propose a new pre-training paradigm, Montage pre-training, which is based only on the detection dataset. Compared with ImageNet pre-training, Montage pre-training takes only $1/4$ computational resources without extra data while achieving on-par or even better performance on the target object detection task. 

Montage pre-training is built upon the observation that a large number of pixels seen by the model during naive training are invalid or less informative, \textit{i.e.}~most pixels/neurons in background regions would not fire during the learning process.
Those excessive background pixels inevitably lead to redundant computational costs.
To tackle this issue, we carefully extract positive and negative samples from original images in the detection dataset for pre-training.
Before being fed into the backbone network, these samples will be assembled in a Montage manner in consideration of their aspect ratios to improve the spatial utilization.
To further improve the pixel level utilization, we design an ERF-adaptive dense classification strategy to leverage the Effective Receptive Field (ERF) via assigning soft labels in the learning objective.
Our Montage pre-training largely takes every pixel seen by the model into account, which greatly reduces the redundancy and provides an efficient and general pre-training solution for object detection.

Our major contributions can be summarized as follows.

(1) We propose an efficient and general pre-training paradigm based only on detection dataset, which eliminates the burdens of additional data.

(2) We design rules of sample extraction, the Montage assembly strategy, and the ERF-adaptive dense classification for efficient pre-training, which largely considers the network utilization and improves the learning efficiency and final performance. 

(3) We validate the effectiveness of our Montage pre-training on various detection frameworks and backbones and demonstrate the versatility of the proposed pre-training strategy. We hope this work would inspire more discussions about the pre-training of object detectors.

%% file: contents/related_works.tex
\noindent \textbf{Classification-based Pre-training for Object Detector.} Recent years have witnessed the significant breakthroughs of deep learning-based object detectors on various scenarios~\cite{girshick2014rich,lin2017focal,ouyang2017chained,redmon2016you,he2017mask,Brazil_2019_ICCV,Ma_2019_ICCV,lu2019grid,Manhardt_2019_CVPR,zhu2019feature,tan2019efficientdet,Ma_2020_ECCV,liu2020deep}. Most of these frameworks follow the standard `pre-training followed by fine-tuning' training procedure, where networks are first pre-trained on the large-scale dataset (\textit{e.g.}~ImageNet~\cite{krizhevsky2012imagenet}) and then fine-tuned on the target detection dataset.
This pre-training paradigm is mainly classification-based and aims to learn strong or universal representations, which speed up the convergence of detection models. Many efforts have been devoted to push the boundary of transferability further through different learning modes such as supervised~\cite{kornblith2019better}, weakly supervised~\cite{mahajan2018exploring,xie2019self}, unsupervised~\cite{he2019momentum} learning, or exploiting larger scale training data such as Instagram-17k~\cite{mahajan2018exploring} and JFT-300M~\cite{sun2017revisiting}. Despite the improvements for transferability, the corresponding expensive training cost of large scale data should not be neglected. Our Montage pre-training is entirely based on detection dataset which eliminates the burden of using external data. Meanwhile, the pre-training process is $4\times$ faster than ImageNet-1k classification training. 

\noindent\textbf{Redundancy in Object Detector.} Sample imbalance is a common source of redundancy for object detection, where many background pixels belonging to easy negative samples contribute no useful information for training. To alleviate this issue, several attempts have been made to improve the efficiency of detection training. OHEM~\cite{shrivastava2016training} tries to solve the imbalance sampling by discarding easy negative samples. Focal loss~\cite{lin2017focal} adopts a weighting factor to reduce loss weight for easy samples. Chen \textit{et al.}~design a more reasonable method for sample evaluation in~\cite{chen2019towards}. Libra R-CNN~\cite{pang2019libra} proposes the IoU-balanced sampling strategy to augment the hard cases. SNIPER~\cite{singh2018sniper} reduces the calculation burden of multi-scale training by only training on selected chips rather than the entire images. All these works mainly focus on the efficiency and performance within detection frameworks, but they still provide inspirations on sample selection in our work. By carefully selecting positive and negative samples for pre-training, the redundancy is significantly reduced, which eventually speeds up the classification pre-training process. 

\noindent \textbf{Object Detector Trained from Scratch.} Many works~\cite{matan1992multi,szegedy2013deep,shen2017dsod,law2018cornernet,li2018detnet,he2019rethinking,zhu2019scratchdet} have proposed another possible training paradigm which is to train the detector from scratch. For instance, DSOD~\cite{shen2017dsod} is motivated by designing a pre-training free detector, but limited to the structure they designed. CornerNet~\cite{law2018cornernet} and DetNet~\cite{li2018detnet} present the results of their models trained from scratch. These efforts indicate that pre-training might be unnecessary when adequate data is available. Furthermore, doubts on ImageNet pre-training are also raised recently.
He~\textit{et al.}~\cite{he2019rethinking} and Zhu~\textit{et al.}~\cite{zhu2019scratchdet} suggest that ImageNet pre-training might be a historical workaround. 
However, although these solutions get rid of the burdens for large-scale external data, the random initialized detection models suffer from the problem of low convergence speed, which comes at the cost of extending training iterations by $4$-$5$ times to obtain competitive models. Inspired by these works, we move steps forward to exploit an efficient pre-training paradigm for pre-training on detection data, which takes the advantages of both fast convergence and no extra data at the same time.

%% file: contents/dataset.tex
\begin{figure}
\centering
\includegraphics[width=1.0\linewidth]{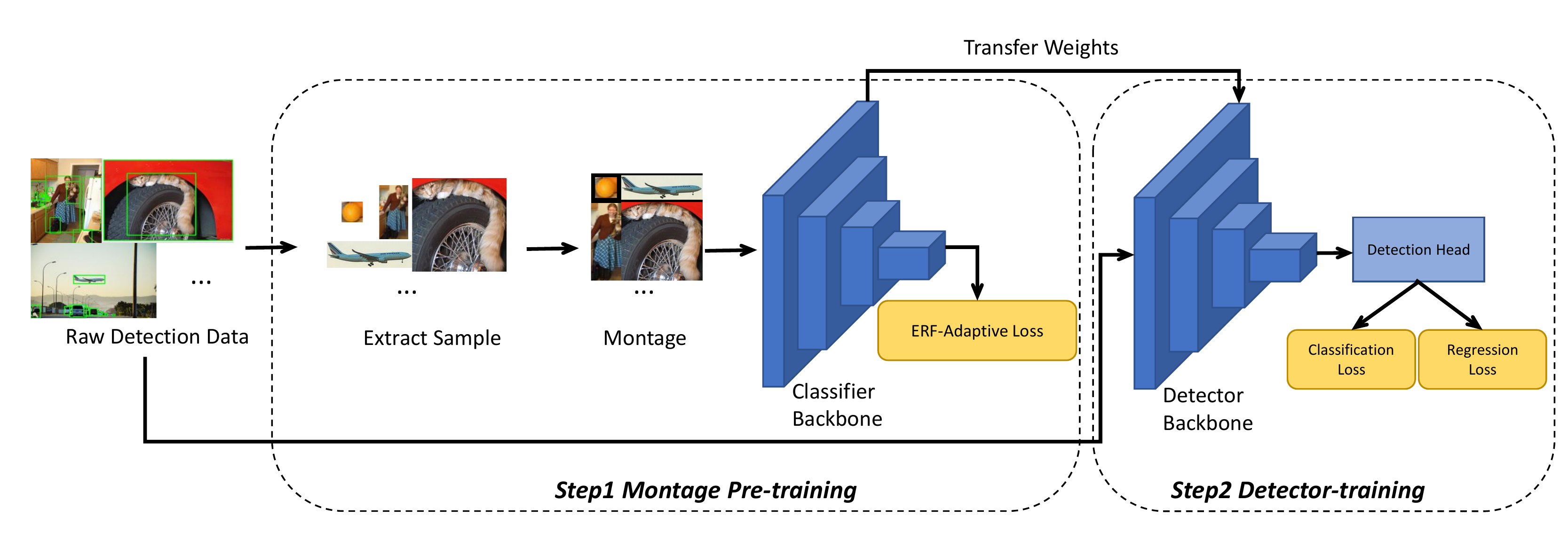}
\caption{Pipeline of the proposed Montage pre-training scheme. Firstly, we will extract positive and negative samples from detection data to build classification dataset. The pre-training process is conducted by Montage, which assembles four objects into a single image, and optimized via ERF-adaptive loss. Finally, the backbones will be fine-tuned on target detection task}
\label{entire_pipeline}
\end{figure}

The pipeline of using the proposed Montage pre-training scheme is shown in Fig.~\ref{entire_pipeline}.
Given a detection dataset $\mathcal{D}$, positive and negative samples will be extracted from the images of $\mathcal{D}$ and saved as classification dataset beforehand (Sec.~\ref{chips_extract}).
These samples will be assembled in a Montage manner (Sec.~\ref{Montage_assembly}) and fed into the detector backbone for pre-training, where an ERF-adaptive loss is used as the loss function (Sec.~\ref{erf_adaptive}).
After pre-training, the object detector will be fine-tuned on $\mathcal{D}$ under the detection task.
Note that our pre-training scheme is flexible and can be applied to object detectors with diverse detection head and backbone architectures.

\subsection{Sample Selection}
\label{chips_extract}
As demonstrated in previous works~\cite{shrivastava2016training,pang2019libra}, balanced sample selection is critical during the training of object detectors.
For efficient pre-training, we carefully select regions extracted from original images as positive and negative samples, which will be further assembled and fed into the detector backbone.

The positive samples are regions that should be classified as one of the $C$ foreground categories in the detection dataset, while the negative samples are background regions.
To effectively select diverse and important samples, we set up following rules for the sample extraction.  (1) For \textbf{positive samples}, we extract regions from the original images according to the ground-truth bounding boxes. The bounding boxes will be randomly enlarged to involve more context information, which is under the consideration that contextual information is beneficial to learn better feature representations~\cite{divvala2009empirical,zheng2011quantifying}. (2) \textbf{Negative samples} are proposals randomly generated from the background regions. To avoid ambiguity, we require that all negative samples meet the requirement $IoU\left(pos, neg\right) = 0$, where $IoU$ indicates Intersection-over-Union. In our pre-training experiments, the ratio of the number of positive samples to negative ones is $10:1$. More details can be found in Section A of the supplementary material.

%% file: contents/jigsaw.tex
\subsection{Montage Assembly}
\label{Montage_assembly}
There are different ways to assemble samples and feed them into the backbone for pre-training.
Two straightforward assembling methods are warping (method 1) or padding (method 2) a sample to a pre-defined input size, \textit{e.g.} ~$224\times224$.
However, forcing all samples to be warped to the same size may destroy the texture information and distort the original shapes, while padding would introduce many uninformative padded pixels and hence bring additional costs in both training time and computational resources.
These two straightforward methods are either harmful or wasteful for the pre-training process.
For more efficient pre-training, we propose to assemble samples in a Montage manner in consideration of the scale and aspect ratio of objects.
Specifically, four samples will be stitched into a new image and then taken as input for pre-training.

\begin{figure}
\centering
\subfigure[Warping]{
  \centering
  \label{single_warp}
  \includegraphics[width=0.25\linewidth]{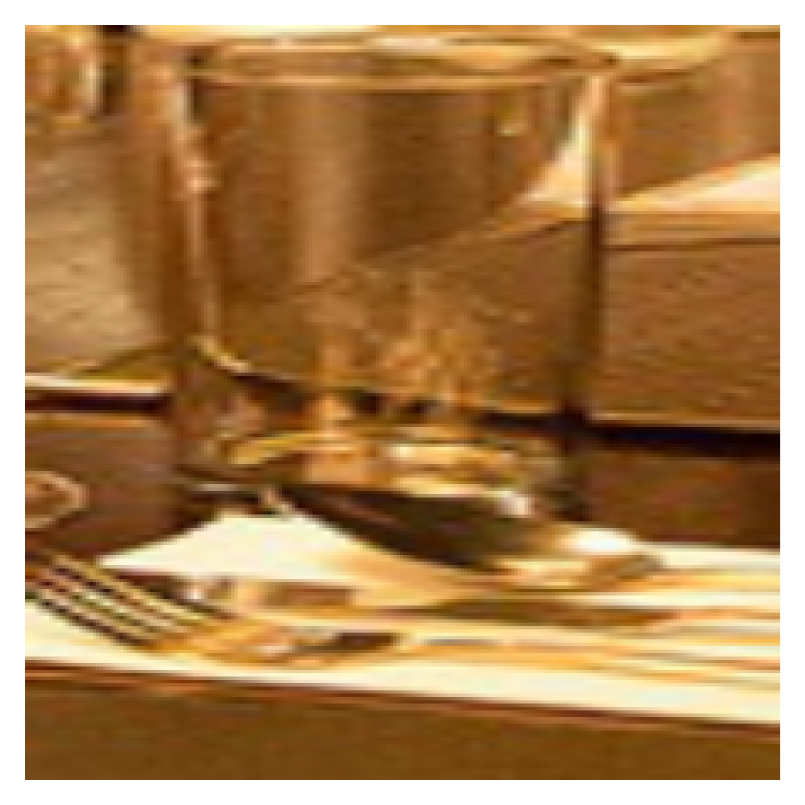}}
\subfigure[Padding]{
  \centering
  \label{iteration}
  \includegraphics[width=0.25\linewidth]{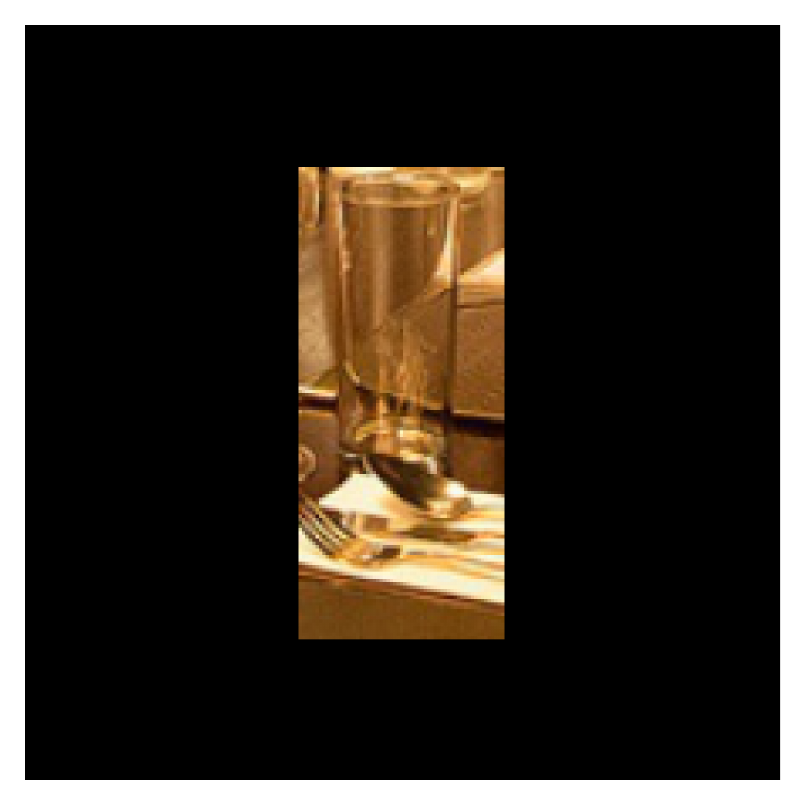}}
 \subfigure[Montage]{
  \centering
  \label{iteration}
  \includegraphics[width=0.25\linewidth]{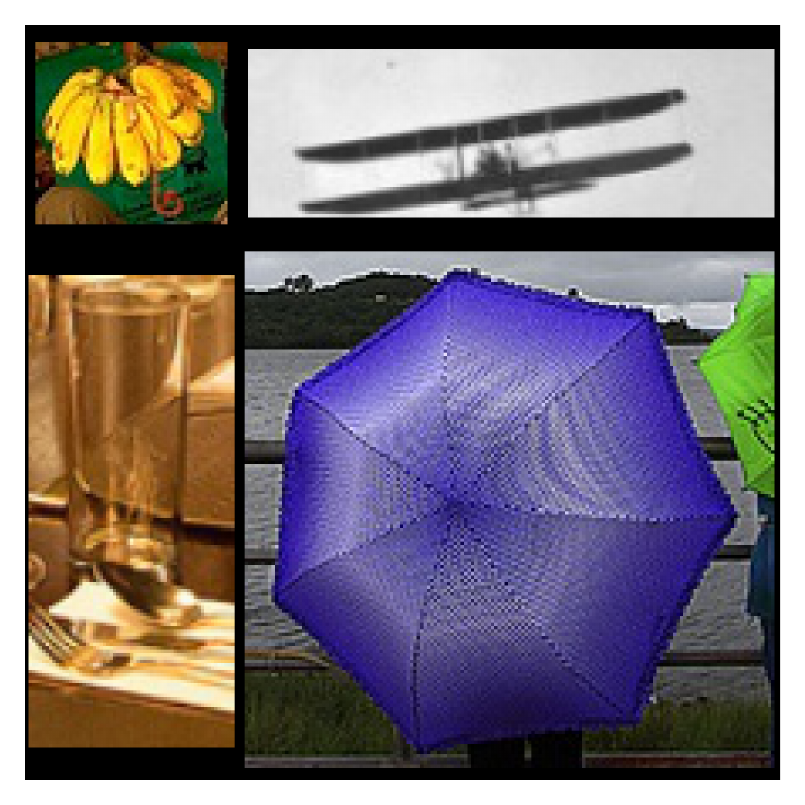}}
\caption{Different methods to adjust sample to pre-defined input size. (a) Warping distorts the original shape or texture. (b) Padding introduces many uninformative pixels. (c) Montage preserves original information while improving space utilization}
\label{assemble_imgs}
\end{figure}

As depicted in Fig.~\ref{assemble_imgs}, compared to warping and padding, our Montage assembly can not only preserve original texture information but also eliminate the uninformative padded pixels.

Objects vary in aspect ratio. Montage assembly takes this property into consideration so that samples could be stitched together more naturally according to their aspect ratios.
To this end, samples will be first divided into three Groups according to their aspect ratios, \textit{i.e.}~Group S (square), T (tall), and W (wide).
Samples in Group S should have the aspect ratios between 0.5 and 1.5, while samples in Group T and W should respectively have aspect ratio smaller than 0.5 and larger than 1.5. For simplicity, samples from Group S, T, and W are referred to as S-samples, T-samples, and W-samples, respectively.

\begin{figure}
    \centering
    \includegraphics[width=0.85\linewidth]{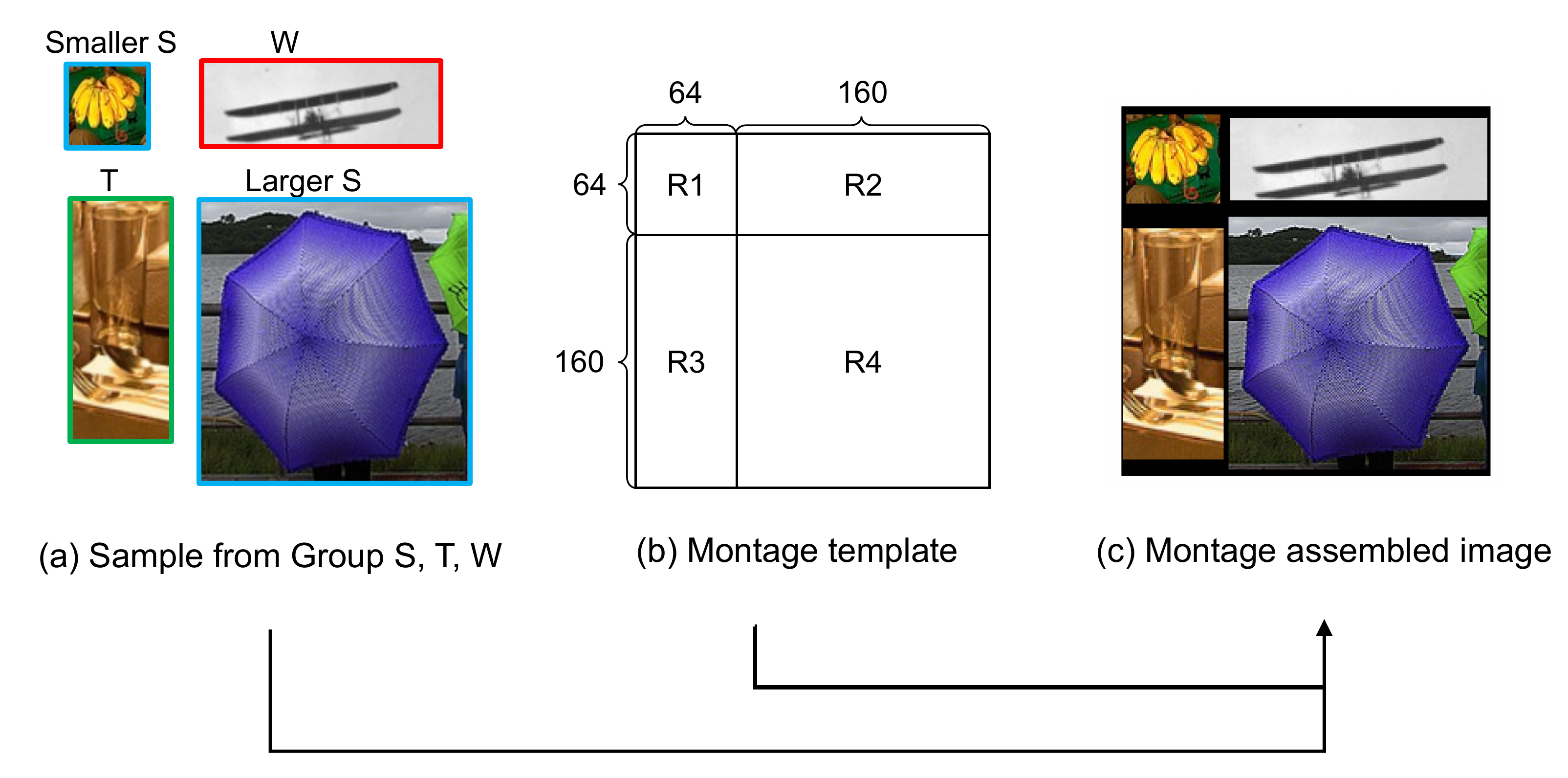}
    \caption{Pipeline for Montage assembled image generation. We first randomly select 2 S-samples, 1 T-sample and 1 W-sample, respectively as shown in (a), then assemble the samples according to the template (b) and get an assembled image (c). The numbers on Montage template denote the height/width of each region}
    \label{gen_montage_pipeline}
\end{figure}

As shown in Fig.~\ref{gen_montage_pipeline}, for every Montage assembled image, 2 S-samples, 1 T-sample, and 1 W-sample will be selected randomly from above three groups and stitched into four regions accordingly. Specifically, the S-sample with smaller bounding box area is at the top-left region, while the larger S-sample is at the bottom-right region.
The T-sample and W-sample will be respectively assigned to bottom-left and top-right regions. Details about sample size adjustment (to fit the template) can be found in Section B of the supplementary material.

%% file: contents/losses.tex
\subsection{ERF-adaptive Dense Classification}
\label{erf_adaptive}

During pre-training, the Montage assembled images will be fed into the backbone network to obtain the feature maps $\mathbf{X}\in \mathbb{R}^{C \times \alpha H \times \alpha W}$ before the final average pooling.
Here we omit the number of samples in $\mathbf{X}$ for simplicity.
Compared to the conventional classification pre-training, Montage pre-training should have different learning strategy since there are four samples stitched in one assembled image. In the following, we discuss two alternative strategies and then introduce our proposed ERF-adaptive Dense Classification.

\noindent\textbf{Global classification.} As shown in Fig.~\ref{gen_montage_pipeline}, an image contains four objects in our Montage assembled image. As an intuitive strategy, we can assign the whole image a single global label, which is the weighted sum of the labels of the four objects according to their region areas.
This strategy could be reminiscent of the CutMix~\cite{Yun_2019_ICCV}, where certain region of the original image will be replaced by a patch from another image and the corresponding label will also be mixed proportionally with the label of the new patch. The visualization of global classification will be provided in the supplementary material. 

\begin{figure}
\centering
\subfigure[Global]{
\label{vis_erf_cutmix}
\centering
\includegraphics[width=0.23\linewidth]{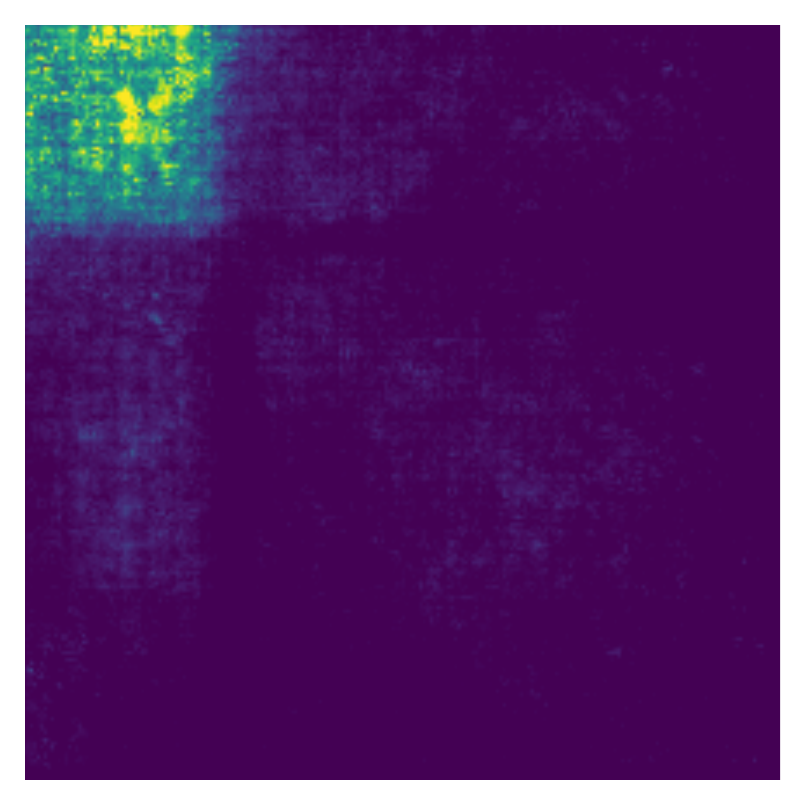}}
\subfigure[Block-wise]{
\label{vis_erf_block}
\centering
\includegraphics[width=0.23\linewidth]{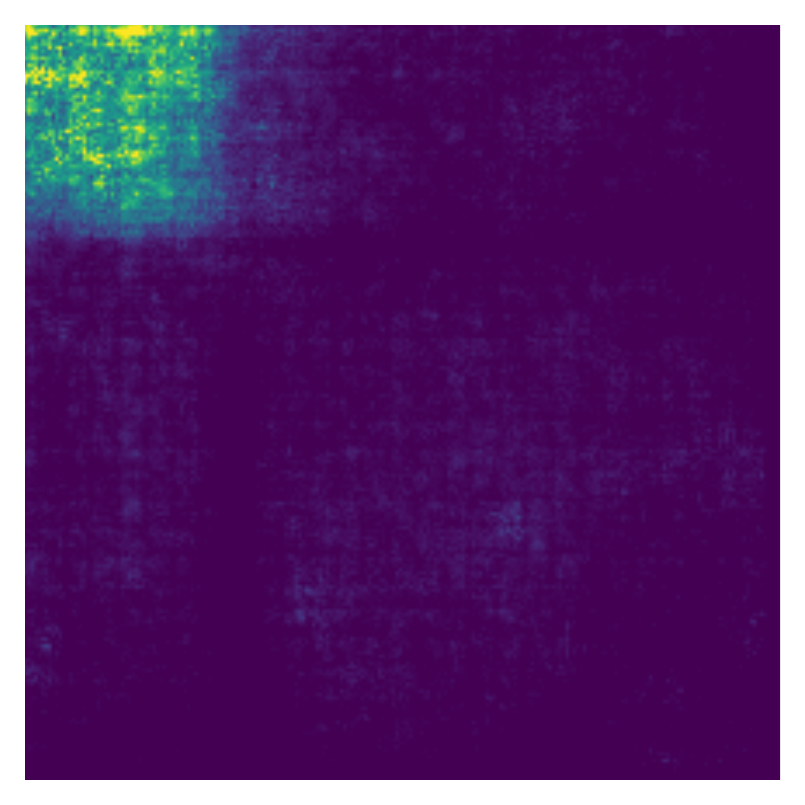}}
\subfigure[ERF-adaptive]{
\label{vis_erf_erf}
\centering
\includegraphics[width=0.23\linewidth]{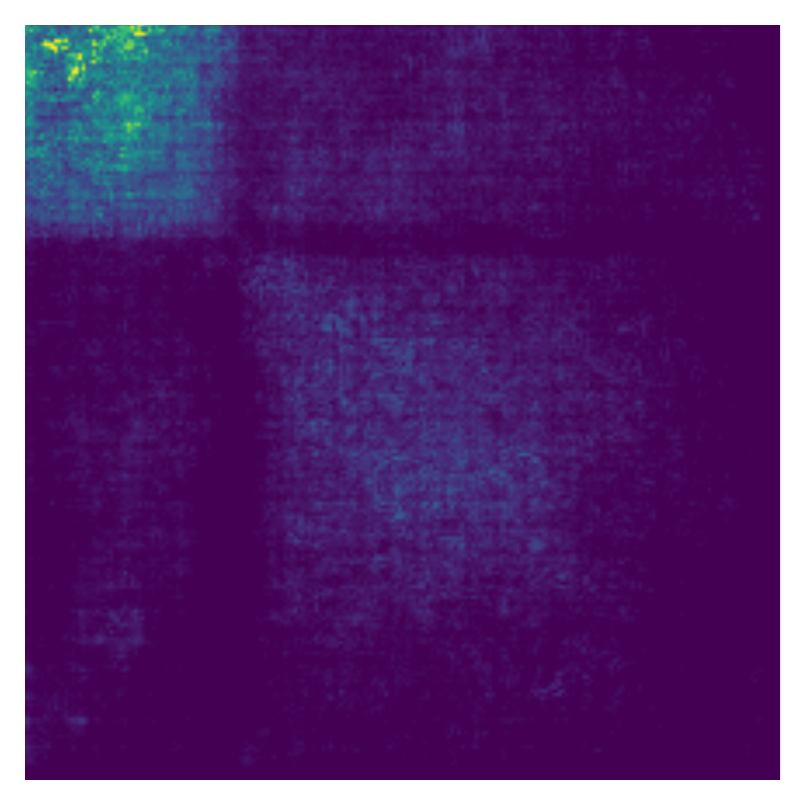}}
\subfigure[ImageNet]{
\label{vis_erf_imagenet}
\centering
\includegraphics[width=0.23\linewidth]{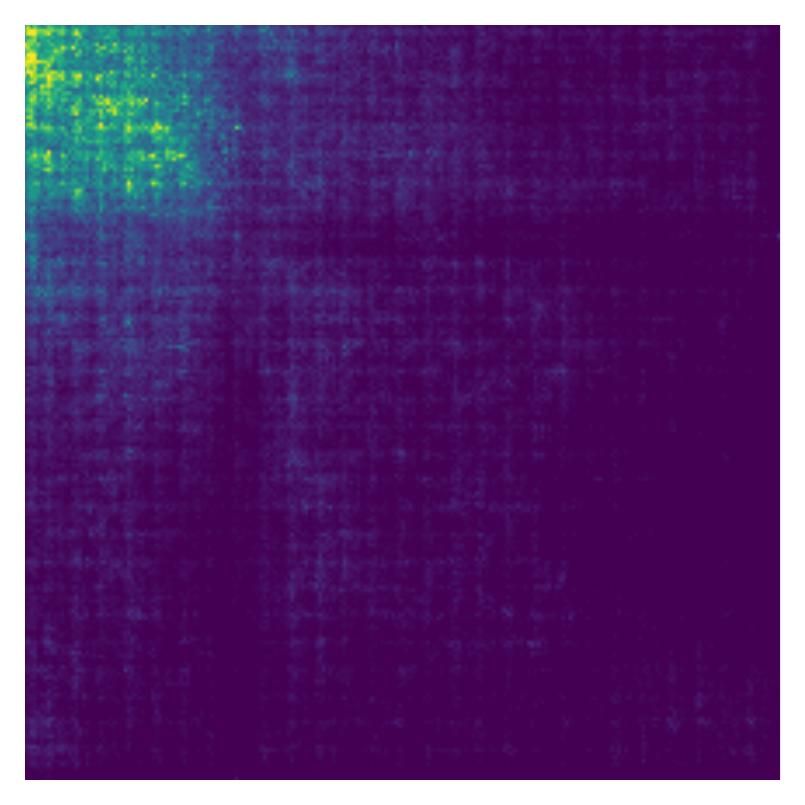}}
\caption{Visualization of Effective Receptive Field of the top-left region for different pre-trained models. (a) Global represents conducting global average pooling and set global label as weighted sum of labels from four regions. (b) Block-wise refers to the intuitive strategy which performs classification individually on each region. (c) ERF-adaptive refers to adopting ERF-adaptive dense classification. (d) ImageNet stands for the ImageNet pre-trained model officially provided by PyTorch~\cite{paszke2019pytorch}}
\label{vis_erf}
\end{figure}

\noindent\textbf{Block-wise classification.} Another intuitive strategy would perform  individually for each block/region, that is, the average pooling is independently applied to the four blocks of feature maps $\mathbf{X}$ corresponding to four samples, followed by individual classification according to the label of each sample.
However, these two intuitive strategies confine the learning of each block to the corresponding sample.
As can be seen in Fig.~\ref{vis_erf_cutmix} and~\ref{vis_erf_block}, the Effective Receptive Field (ERF)~\cite{luo2016understanding} of the top-left region in $\mathbf{X}$ mainly concentrates on the area of the corresponding smaller S-sample. The confined receptive field may empircally degrade the performance of deep models, as illustrated in~\cite{liu2018receptive,li2019scale,peng2019efficient}. The visualization of block-wise classification will be provided in the supplementary material. 

\begin{figure}
    \centering
    \includegraphics[width=0.7\linewidth]{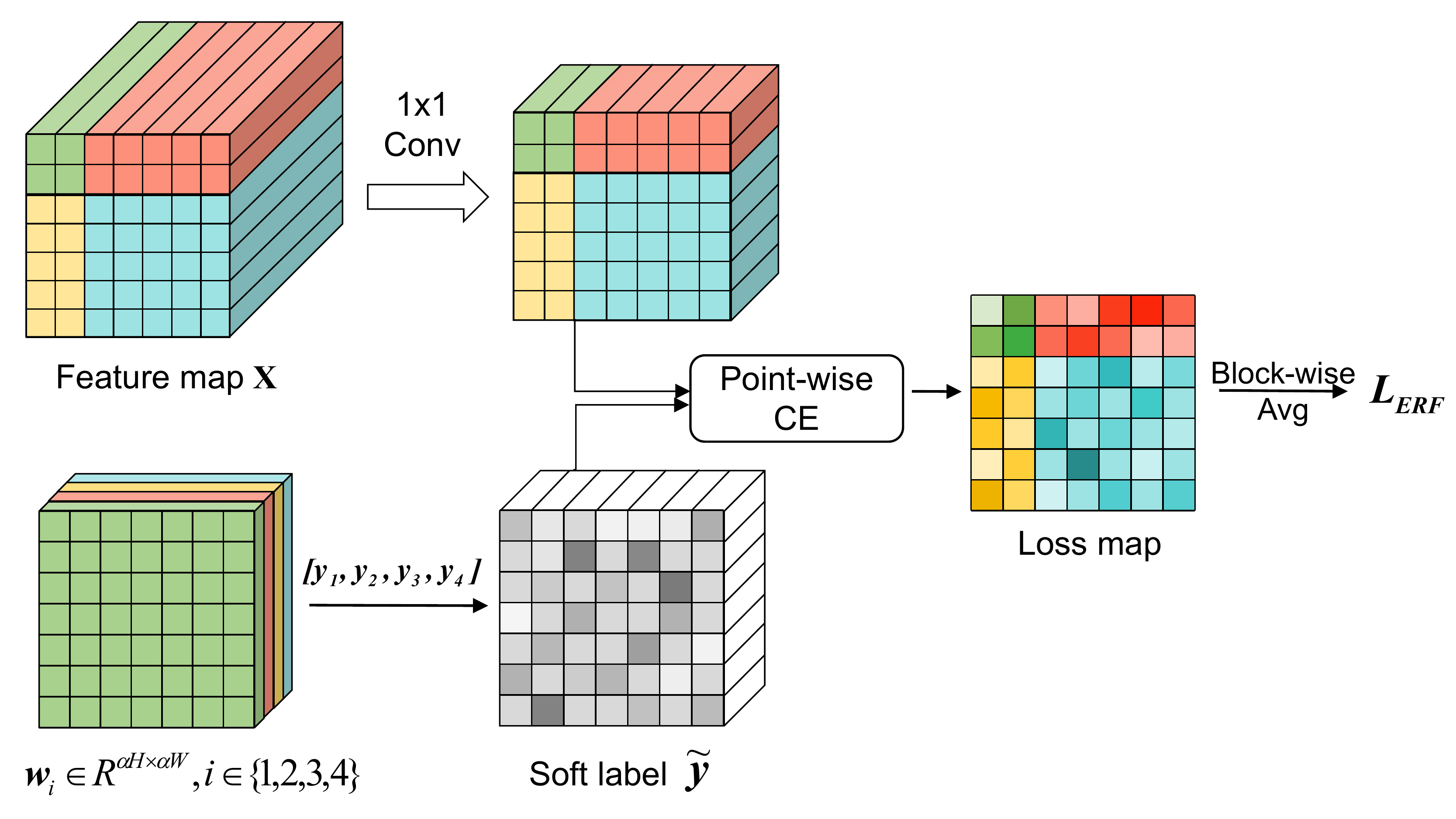}
    \caption{Process of our Dense Classification Strategy. We use different colors to distinguish different regions, \textit{e.g.}~green for $R_1$, and the brightness difference in soft label and loss map represents different values. The feature map $\mathbf{X}$ is convolved by a $1\times1$ kernel to reduce the number of channels to $C$ (category number). Given the weight $w_i$ of label $\mathbf{y_i}$, we obtain the soft label tensor for each point. Then the cross-entropy loss is densely imposed on the feature map and we will get the loss value at each point (denoted as loss map). In the mean time, the loss weight for a position is calculated according to the ERF of the position. After that, block-wise average is exerted on the loss map to generate average losses for each region. The final ERF-adaptive loss is the mean of four region losses. Best viewed in color}
    \label{dense_cls}
\end{figure}

\noindent\textbf{Our strategy.} To largely take every seen pixel into account, we propose an ERF-adaptive Dense Classification strategy to perform classification for each position at $\mathbf{X}$, where its soft labels are computed based on the corresponding effective receptive field. The process is depicted in Fig.~\ref{dense_cls}.

Specifically, for the four regions in the Montage template as shown in Fig.~\ref{gen_montage_pipeline}(b), we denote $\mathbf{y}_i$ as the original label for the region $R_i$, $i=1, 2, 3, 4$.

At the position $(j, k)$ of feature map $\mathbf{X}$ ($j=1, \ldots, \alpha H, k=1, \ldots, \alpha W$), the soft label $\tilde{\mathbf{y}}^{j,k}$ is the weighted sum of four labels:
\begin{equation}
    \tilde{\mathbf{y}}^{j,k} = \sum_{i=1}^4 w_{i}^{j,k}\mathbf{y}_i,
\label{soft_label}
\end{equation}

where the weight $w_{i}^{j,k}$ is dependent on its ERF. At the position $(j, k)$ of feature map $\mathbf{X}$ ($j=1, \ldots, \alpha H, k=1, \ldots, \alpha W$), we obtain the corresponding ERF $\mathbf{G}^{j,k}\in\mathbb{R}^{H \times W}$ on the input space.
Then, the weight $w^{j,k}_{i}$ for the label $\mathbf{y}_i$ at position $(j, k)$ should be proportional to the ratio of the summed activation within the region $R_i$ to the whole summed activation.
Moreover, if the position $(j, k)$ is in region $R_i$, we empirically set a threshold $\tau$ for $w^{j,k}_{i}$ to make sure that  $\mathbf{y}_i$ is dominant at region $R_i$.
Hence, for the position $(j, k)$ at region $R_r$, we have the weight $w^{j,k}_{i}$ of the label $\mathbf{y}_i$ as follows:

\begin{equation}
w^{j,k}_{i} = 
\left\{ 
\begin{array}{l}
\max(\tau, \frac{\sum_{h=1,w=1}^{H,W} {g}^{j,k}_{h,w} \cdot {m}^i_{h,w}}{\sum_{h=1,w=1}^{H,W} {g}^{j,k}_{h,w}}), \textrm{if}~i = r,\\
(1 - w^{j,k}_{r}) \frac{\sum_{h=1,w=1}^{H,W} {g}^{j,k}_{h,w} \cdot {m}^i_{h,w}}{\sum_{h=1,w=1}^{H,W} {g}^{j,k}_{h,w} \cdot (1-{m}^r_{h,w})}, \textrm{if}~i \neq r, \\
\end{array}
\right.
\label{ori_label_weight}
\end{equation}

where ${g}^{j,k}_{h,w}$ is the element at position $(h, w)$ on the corresponding ERF matrix $\mathbf{G^{j, k}}$, $m^i_{h,w}$ refers to the value at position $(h, w)$ on binary mask $\mathbf{M}^i\in\{0,1\}^{H\times W}$. The binary mask $\mathbf{M}^i$ is used to select the region $R_i$ in ERF.

Denote $\mathbf{x}^{j, k}\in \mathbb{R}^{C}$ as the features at the position $(j, k)$ of $\mathbf{X}$ ($j=1, \ldots, \alpha H, k=1, \ldots, \alpha W$).
After obtaining the weights $\{w^{j,k}_{i}\}_{i=1}^{4}$, we perform dense classification upon the feature $\mathbf{x}^{j, k}$, where its soft label $\tilde{\mathbf{y}}^{j,k}$ is defined in Eq.~(\ref{soft_label}).
In our implementation, the final fully connected layer is replaced by a $1\times1$ convolution layer and the cross-entropy loss is imposed on the category prediction at every position. To make a balance among different regions, the final ERF-adaptive loss is the block-wise average of the loss map, as the last step in Fig.~\ref{dense_cls}. We also need to clarify that the weights of soft label Eq.~(\ref{soft_label}) are updated at every 5k iterations instead of at each iteration. Thus, even if dense classification is adopted, its effect on training time is negligible. Correspondingly, since ERF will be updated regularly during the whole training process, different initialization choices of ERF will not affect the final results. We choose the method to calculate ERF based on the randomly initialized network parameters.

The effective receptive field of the top-left region for the different pre-training strategies is visualized in Fig.~\ref{vis_erf}.
Our strategy in Fig.~\ref{vis_erf_erf} has the largest ERF among the above three strategies.

\noindent\textbf{Relationships among Different Strategies.}
The above three strategies perform classification at different scale levels, where the proposed ERF-adaptive classification is the most fine-grained one while the global classification is the coarsest one.
Compared with the other two alternative strategies, the proposed one has different soft labels for each position at $\mathbf{X}$.
The ERF-adaptive dense classification would be equivalent to the block-wise classification with threshold $\tau$ set to 1. The block-wise classification would be also equivalent to the global classification if the region losses are re-weighted in a CutMix manner. Under different label assignment strategies, the pre-trained model has different pixel level utilization and hence behaves differently. Comparison of performance for the strategies can be found in the supplementary material. 

%% file: contents/experiments.tex
\subsection{Implementation details}

This section introduces the implementation details of the classification pre-training. Details of data augmentation in pre-training and detector training settings will be provided in the supplementary.

Unless otherwise specified, the models are pre-trained for 64k iterations on 8 Tesla V100 GPUs with the total batch size of 512. Note that the batch size 512 is for Montage assembled images, so the total number of individual samples in each batch is 2048 (an assembled image consists of 4 samples). Warm-up is used during the first 1250 iterations, where the learning rate starts from 0.2 and then linearly increases to 0.8. Afterwards, the learning rate decreases to 0.0 following a cosine scheduler. The weight decay is $1\mathrm{e}{-4}$. We update weights $w^{j,k}_{i}$ of soft labels in Eq.~\ref{ori_label_weight} for every 5k iterations. The threshold $\tau$ in Eq.~\ref{ori_label_weight} for ERF-adaptive classification is set to 0.7. The data augmentation implementation can be found in the supplementary. 

\subsection{Main Results}

We conduct the Montage pre-training process based on samples extracted from MS-COCO train2017 split, and fine-tune the detection models on the same dataset. The backbone is ResNet-50. Note that the Montage pre-training process only consumes \textbf{1/4} computation resources compared with ImageNet pre-training. As reported in Table~\ref{frame_result}, the results show that the models using our Montage pre-training strategy are able to achieve on-par or even better performances compared with the ImageNet pre-training counterparts for various detection frameworks.  For original Faster R-CNN~\cite{ren2015faster}, the AP increases from 34.8\% to 36.3\% ($+1.5\%$), for Faster R-CNN with FPN~\cite{lin2017feature}, AP increases from 36.2\% to 36.5\% ($+0.3\%$), for Mask R-CNN with FPN~\cite{he2017mask}, AP increases from $37.2\%$ to $37.4\%$($+0.2\%$). 

We notice that the improvement is most significant in the original Faster R-CNN structure (denoted as C4 in Table~\ref{frame_result}). We suspect the possible reason is that, compared with FPN structure, the backbone accounts for a larger proportion in C4. In other words, for detection models with FPN structure, the lateral connections and entire structures at the second stage will be randomly initialized without being transferred from pre-trained model. But for the original Faster R-CNN, the main part of the second stage is still transferred from the pre-trained network. We speculate that the improvement of detection models with FPN may be consistent with that of C4 if the FPN structure is incorporated into pre-training, and we will leave this exploration for future work.

\setlength{\tabcolsep}{4pt}
\begin{table}
\begin{center}
\caption{Results on different detection frameworks with backbone ResNet-50. The cost refers to pre-training cost and the unit is GPU days. The AP results are evaluated on COCO val2017. `C4' denotes original Faster R-CNN without FPN~\cite{ren2015faster}, `FPN' denotes Faster R-CNN with FPN~\cite{lin2017feature}, `Mask' denotes Mask R-CNN with FPN \cite{he2017mask}. `+~ImageNet' means the backbone is pre-trained on ImageNet dataset. `+~Montage' denotes that the backbone is pre-trained with our Montage strategy. $\Delta$ measures the difference in absolute AP or cost between adopting Montage and ImageNet pre-trained backbones, respectively}
\label{frame_result}
\begin{tabular}{llllllll}
\hline
Method & Cost & AP & AP$_{50}$ & AP$_{75}$ & AP$_s$ & AP$_m$ & AP$_l$ \\
\hline
C4\cite{ren2015faster} + ImageNet & 6.80 & 34.8 & 55.5 & 36.8 & 18.3 & 38.7 & 48.4 \\
C4\cite{ren2015faster} + Montage & 1.73 & 36.3 & 56.5 & 38.9 & 18.9 & 40.8 & 49.7 \\
$\Delta$ & \gre{-5.07} & \inc{+1.5} & \inc{+1.0} & \inc{+2.1} & \inc{+0.6} & \inc{+2.1} & \inc{+1.3} \\
\hline
FPN\cite{lin2017feature} + ImageNet & 6.80 & 36.2 & 58.0 & 39.2 & 21.2 & 39.9 & 45.6 \\
FPN\cite{lin2017feature} + Montage & 1.73 &  36.5 & 58.3 & 39.2 & 22.2 & 40.4 & 45.8 \\
$\Delta$ & \gre{-5.07} & \inc{+0.3} & \inc{+0.3} & \inc{0.0} & \inc{+1.0} & \inc{+0.5} & \inc{+0.2} \\
\hline
Mask\cite{he2017mask} + ImageNet & 6.80 & 37.3 & 59.0 & 40.3 & 21.9 & 40.6 & 46.2 \\
Mask\cite{he2017mask} + Montage & 1.73 & 37.5 & 58.9 & 40.6 & 22.8 & 41.2 & 46.9 \\
$\Delta$ & \gre{-5.07} & \inc{+0.2} & \dec{-0.1} & \inc{+0.3} & \inc{+0.9} & \inc{+0.6} & \inc{+0.7} \\
\hline
\end{tabular}
\end{center}
\end{table}
\setlength{\tabcolsep}{1.6pt}

\subsection{Ablation study}

\begin{figure}
\subfigure[]{
  \centering
  \label{erf_thr}
  \includegraphics[width=0.45\linewidth]{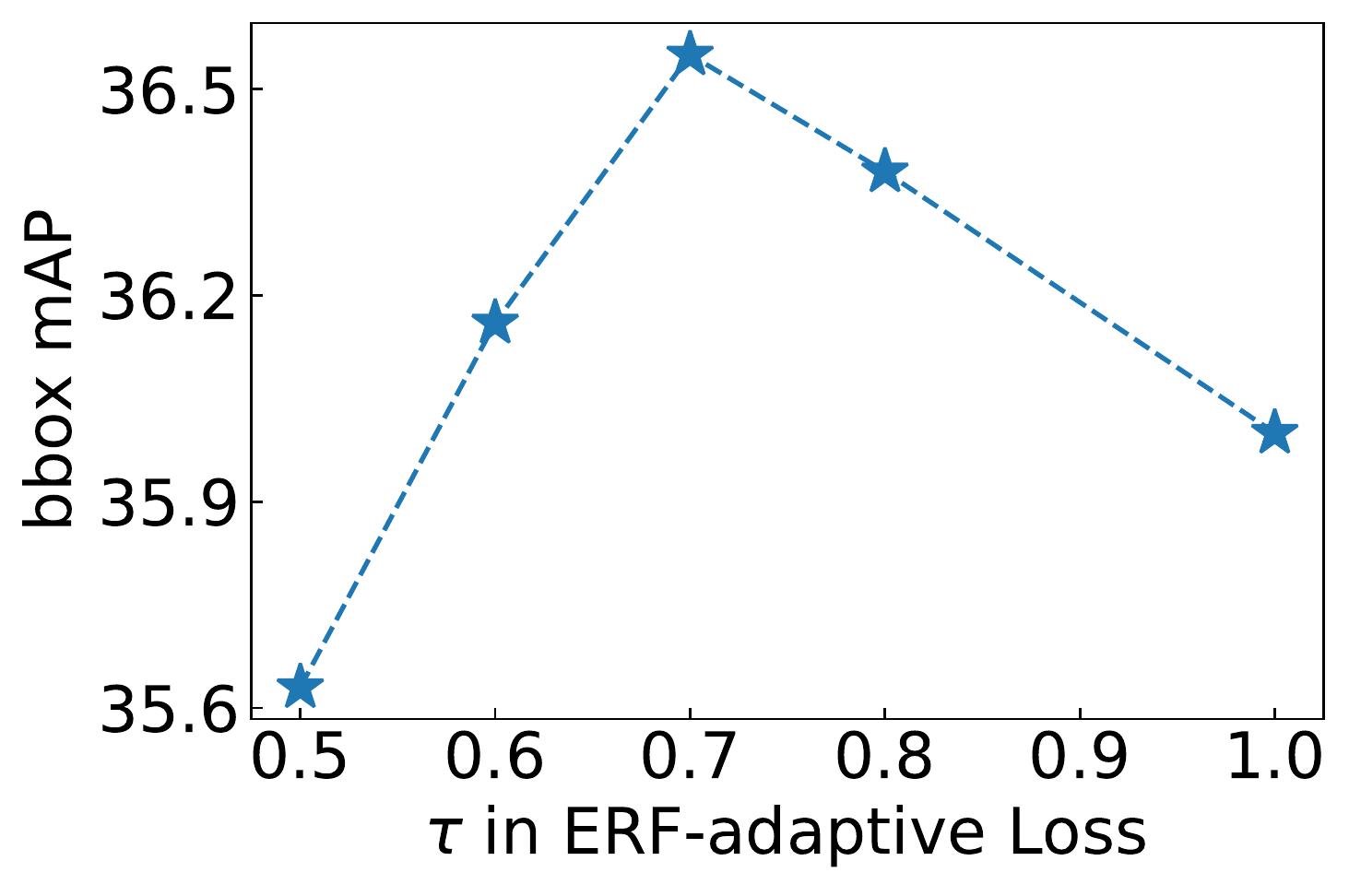}}
\subfigure[]{
  \centering
  \label{iteration}
  \includegraphics[width=0.45\linewidth]{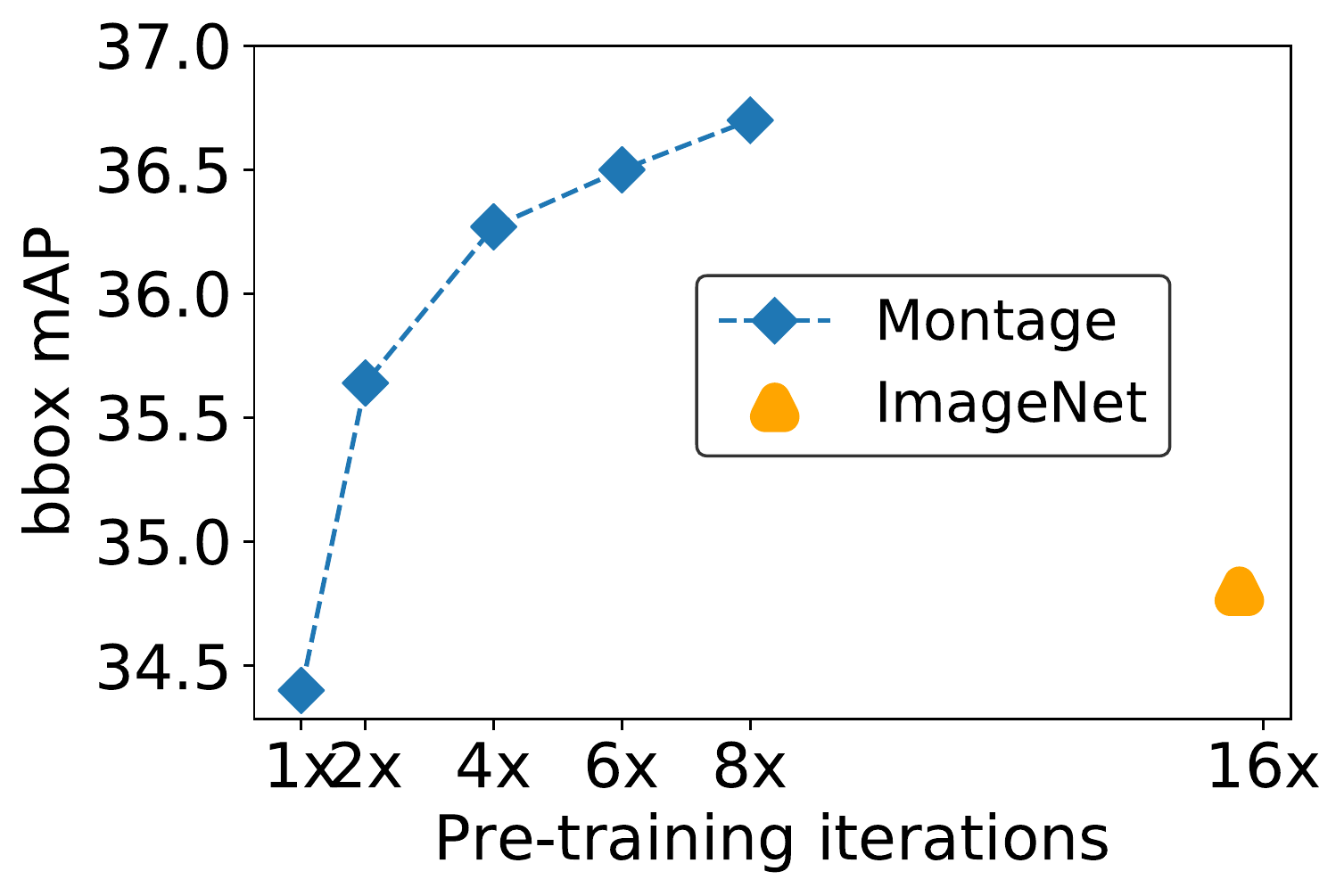}}
\caption{(a) Trend of bbox mAP with different $\tau$ settings in ERF-adaptive loss. (b) bbox mAP for different pre-training iterations, the values on $x$-axis stand for multiple of 16k iterations. Our pre-training approach uses $4\times$ as the default setting, which requires $1/4$ the number of iterations but achieves 1.5\% higher mAP when compared with ImageNet counterpart. The results are evaluated on COCO val2017}
\label{ablation}
\end{figure}

\textbf{Threshold for ERF-adaptive Dense Classification.} When  ERF-adaptive Dense Classification is used, there is a threshold $\tau$ in Eq.~(\ref{soft_label}) to make sure that the original label $y_i$ is dominant at its corresponding region $i$. We explore the effects of this threshold and the results are depicted in Fig.~\ref{erf_thr}. Although using mixed labels is beneficial, relatively low proportion of original label (\textit{e.g.}\ 0.5) may still hinder the pre-training. As the threshold becomes higher, the loss is gradually approaching the use of single hard label for each point, which may suffer from relatively confined receptive field, as analyzed in Section~\ref{erf_adaptive}. Therefore, it is important to choose proper threshold and we find 0.7 is an ideal choice. Fig.~\ref{erf_thr} also shows that setting the threshold in [0.6 0.8] will not cause much variation in mAP. Therefore, the experimental results are not so sensitive to this hyper-parameter.

\noindent\textbf{Iterations for pre-training.} We also investigate the influences of changing the pre-training iterations and visualize the results in Fig.~\ref{iteration}. Naturally, increasing training iterations will provide better pre-trained models, which leads to better detection performance. But we also observe that the gains from longer iterations are not so significant after 64k iterations (4$\times$ in Fig.~\ref{iteration}). Considering the trade-off between performance and computation, we choose to train 64k iterations during pre-training, which consumes only $1/4$ computation resources but achieve 1.5\% higher mAP compared with ImageNet pre-training. 

\noindent\textbf{Different backbone structures.} We also implement our pre-training strategy on different backbone structures to evaluate the versatility. The results in Table \ref{backbones} show that Montage pre-training strategy does not rely on specific network structures but will consistently keep on-par performance or obtain improvements.

\setlength{\tabcolsep}{4pt}
\begin{table}
\begin{center}
\caption{Results for different backbone structures evaluated on COCO val2017. The detection framework is original Faster R-CNN. ImageNet means the backbone is trained on ImageNet dataset. Montage denotes that the backbone is pre-trained with Montage strategy. X101-32x4d refers to ResNeXt101-32x4d~\cite{xie2017aggregated}}
\label{backbones}
\begin{tabular}{lllllll}
\hline
Method & AP & AP$_{50}$ & AP$_{75}$ & AP$_s$ & AP$_m$ & AP$_l$ \\
\hline
ResNet-101 + ImageNet & 38.3 & 58.9 & 41.1 & 20.0 & 42.8 & 53.0 \\
ResNet-101 + Montage & 39.2 & 59.6 & 42.0 & 20.6 & 43.3 & 54.8 \\
$\Delta$ & \inc{+0.9} & \inc{+0.7} & \inc{+0.9} & \inc{+0.6} & \inc{+0.5} & \inc{+1.8} \\ 
\hline
X101-32x4d + ImageNet & 40.2 & 61.2 & 43.2 & 21.2 & 44.6 & 55.7 \\
X101-32x4d + Montage & 40.2 & 61.0 & 43.0 & 21.3 & 44.5 & 55.7 \\
$\Delta$ & \inc{0.0} & \dec{-0.2} & \dec{-0.2} & \inc{+0.1} & \dec{-0.1} & \inc{0.0} \\
\hline
\end{tabular}
\end{center}
\end{table}
\setlength{\tabcolsep}{1.6pt}

\subsection{Compatibility to other designs}

\begin{figure}
    \centering
    \includegraphics[width=0.75\linewidth]{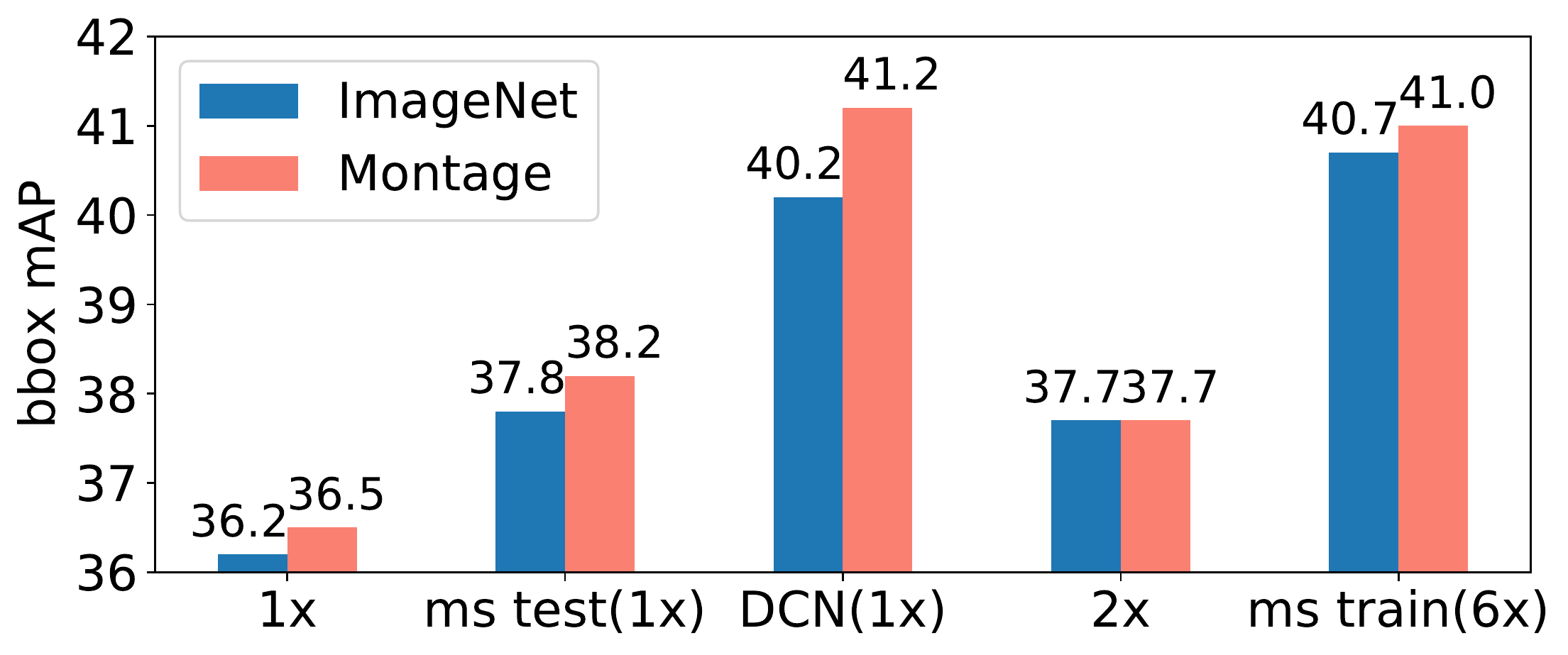}
    \caption{Comparison between \textbf{ImageNet} pre-training and \textbf{Montage} pre-training of various strategies on Faster R-CNN FPN framework with ResNet-50 backbone, the results are evaluated on COCO val2017. Strategies include: (1) $1\times$: serving as original strategy that train for $1\times$ schedule (13 epochs), (2) ms test: adding multi-scale augmentation during test stage, (3) DCN: replace the $3\times3$ convolution layers of stage 2-4 in backbone with $3\times3$ deformable convolution layer \cite{dai2017deformable}, (4) $2\times$: extending the training time to $2\times$ schedule, (5) ms train: implementing multi-scale augmentation during train and test stage and extending training epochs to $6\times$ schedule}
    \label{bells_whistles}
\end{figure}

We also examine the compatibility of Montage pre-training strategy with commonly used designs in object detection, including longer training iterations (2x schedule), deformable convolution \cite{dai2017deformable}, multi-scale augmentation, etc. The results in Fig.~\ref{bells_whistles} indicate that Montage pre-training can still achieve comparable or even higher performance even on various enhanced baselines. 

It is worth noting that the most obvious improvement has been achieved when replacing some convolution layers to deformable convolution layers. We suspect that this improvement may come from the relief of domain shift between pre-training dataset and detection dataset. Therefore, our approach has the potential of further boosting the performance gains from new designs on backbones.

\subsection{Comparison with vanilla training detection from scratch}

We also compare our Montage pre-training strategy with the vanilla training detection from scratch method (denoted as \textbf{vanilla scratch} for simplicity). Vanilla scratch and our strategy share an advantage that the entire training process is only based on detection dataset without introducing any external data. However, adopting pre-training process will speed up the convergence of detection models, which helps the models achieve better performance under common training iterations, such as $1\times$ or $2\times$ schedules. The results are presented in Table~\ref{comp_scratch}. To make a fair comparison, we keep total training costs similar for the two methods, that is, the total costs in our method include both pre-training and detection training consumptions. We follow the experimental settings for vanilla scratch in~\cite{he2019rethinking} where all batch norm layers in the network are replaced by group norm~\cite{wu2018group}. The batch norm layers are frozen at detection stage when transferring from our pre-trained backbones. The results indicate that even with group normalization, which is proven to improve the performance of detection models, vanilla scratch still shows suboptimal performance compared with our pre-training strategy. 

\setlength{\tabcolsep}{4pt}
\begin{table}
\begin{center}
\caption{Comparison of vanilla scratch and Montage pre-training under similar computation costs. The detection framework is Faster R-CNN with backbone ResNet-50, and the AP results are evaluated on COCO val2017. The unit for total cost is GPU days. $1\times$ refer to training detection models for widely adopted $1\times$ schedule, while $2\times$ to extend the iterations to twice}
\label{comp_scratch}
\begin{tabular}{lclll}
\hline
Method & Total Cost & AP & AP$_{50}$ & AP$_{75}$ \\
\hline
Vanilla scratch & 6.0 & 28.6 & 46.5 & 30.1  \\
Montage + $1\times$ & 5.8 & 36.3 & 56.5 & 38.9 \\
\hline
Vanilla scratch & 9.5 & 32.6 & 51.6 & 34.7 \\
Montage + $2\times$ & 9.2 & 37.5 & 57.6 & 40.7 \\
\hline
\end{tabular}
\end{center}
\end{table}
\setlength{\tabcolsep}{1.6pt}

%% file: contents/discussion.tex
The possible reasons for the Montage pre-training being effective are as follows:

First, there is domain gap between the ImageNet dataset and objective detection dataset, such as the data distribution and category. Directly pre-training on the target detection dataset will alleviate the domain gap and help obtain better initialization. However, simply changing the pre-training dataset is not enough and the specially designed pre-training strategy is necessary. Table~\ref{comp_coco_montage} shows the comparison between simply replacing dataset from ImageNet to MS-COCO and Montage pre-training. We can see that pre-training on MS-COCO classification for the same training time as ours (1.73 GPU days) performs worse than ImageNet classification and our approach. Thus, directly training on MS-COCO by saving the training computational costs leads to drop in detection accuracy. If the training time on MS-COCO is extended to the same as that on ImageNet (6.80 GPU days), the final AP will be similar to ImageNet, but still worse than our approach. Therefore, preserving detection accuracy and saving computational cost at the same time cannot be simply brought by adopting MS-COCO classification dataset, but our Montage pre-training strategy is able to preserve detection performance under lower computation costs. 

\setlength{\tabcolsep}{4pt}
\begin{table}
\begin{center}
\caption{Experimental results on ImageNet pre-training, Montage Pre-training using the same and higher computational costs. The detection framework is Faster R-CNN with backbone ResNet50, and the AP results are evaluated on COCO val2017. The unit of cost is GPU days. ImageNet refers to pre-training on ImageNet dataset and MS-COCO represents training on samples extracted from MS-COCO dataset, as illustrated in Section~\ref{chips_extract}}
\label{comp_coco_montage}
\begin{tabular}{lcl}
\hline
Method & Total Cost & AP \\
\hline
ImageNet & 6.80 & 34.8 \\
MS-COCO \textit{w/o} Montage (higher cost) & 6.80 & 34.4 \\
MS-COCO \textit{w/o} Montage & 1.73 & 33.5 \\
MS-COCO \textit{w.} Montage (ours) & 1.73 & 36.3 \\
\hline
\end{tabular}
\end{center}
\end{table}
\setlength{\tabcolsep}{1.6pt}

Second, we speculate that the improved training efficiency of Montage pre-training comes from the reduction of redundancy in training dataset. In the training dataset, the amount of foreground and background pixels are imbalanced, especially for the detection dataset. Thus, we design a reasonable sampling strategy to compose training data, which makes the pre-trained networks focus more on positive samples. By discarding the useless pixels and effectively assembling training samples, our Montage pre-training strategy contributes to the reduction of redundancy, which explains the improvements of training efficiency. By assigning soft labels at the regions that overlap with multiple objects leads, more supervised signals are provided for learning better features.

Finally, the ERF-adaptive loss has positive effects on expanding the effective receptive field of the pre-trained models, which provides stronger supervision signals and obtains pre-trained model more appropriate for the detection task. Larger receptive field helps to promote the performance of detection models, as demonstrated in previous works ~\cite{li2019scale,liu2018receptive,dai2017deformable}.

%% file: contents/conclusion.tex
In this work, we present a choice to obtain cheaper lunch on pre-training for object detection, which is able to reduce the consumption of pre-training  to $1/4$ compared with the original ImageNet pre-training, while achieving on-par or even higher performance. We define a novel pre-training paradigm based only on detection dataset, which eliminates the burdens of extra training data while retaining the advantage of fast convergence. Our efficient Montage Pre-training facilitates training from scratch, which can reduce the computational cost when directly using network compression and neural architecture search~\cite{zhou2020econas,jiang2020sp} for target tasks like object detection. We expect this work would help researchers reduce the trial-and-error cost, inspire more future research on pre-training process, and facilitate new backbone CNN architecture design/search ~\cite{liang2019computation} tailored for object detection. 

%% file: contents/supp.tex
\renewcommand\thefigure{A\arabic{figure}}
\renewcommand\thetable{A\arabic{table}}
\setcounter{figure}{0}
\setcounter{table}{0}

\section{Details about sample generation}
\begin{figure}
    \centering
    \includegraphics[width=1.0\linewidth]{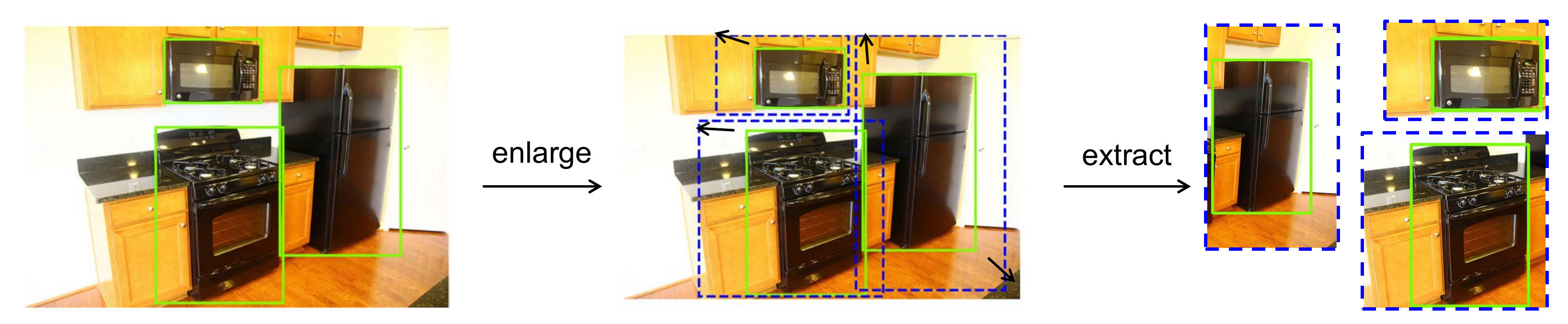}
    \caption{The process for generating positive samples. First, we will use ground-truth bounding boxes to locate target regions. Then the regions are randomly enlarged to incorporate context information. Finally the enlarged regions are extracted from the original images as positive samples for pre-training. The green solid lines stand for GT-bounding boxes and blue dash lines for enlarged boxes}
    \label{gen_pos}
\end{figure}

The positive samples will be randomly enlarged to involve more contextual information, which is beneficial for the training process. Specifically, the upper-left and bottom-right corner points of the ground-truth bounding boxes randomly move outwards so that the width/height could be up to $2\times$ of the original width/height. Enlarged bounding boxes beyond original images will be truncated at the edges. Fig.~\ref{gen_pos} depicts the generation of positive samples. Experimental results indicate that adding context to the positive samples during pre-training will bring 1.6\% mAP gain for Faster R-CNN~\cite{ren2015faster} with ResNet-50 backbone.

\section{Sample adjustment strategy for Montage assemble process}
\begin{figure}
\centering
\subfigure[Samples]{
\label{samples_sup}
\centering
\includegraphics[width=0.63 \linewidth]{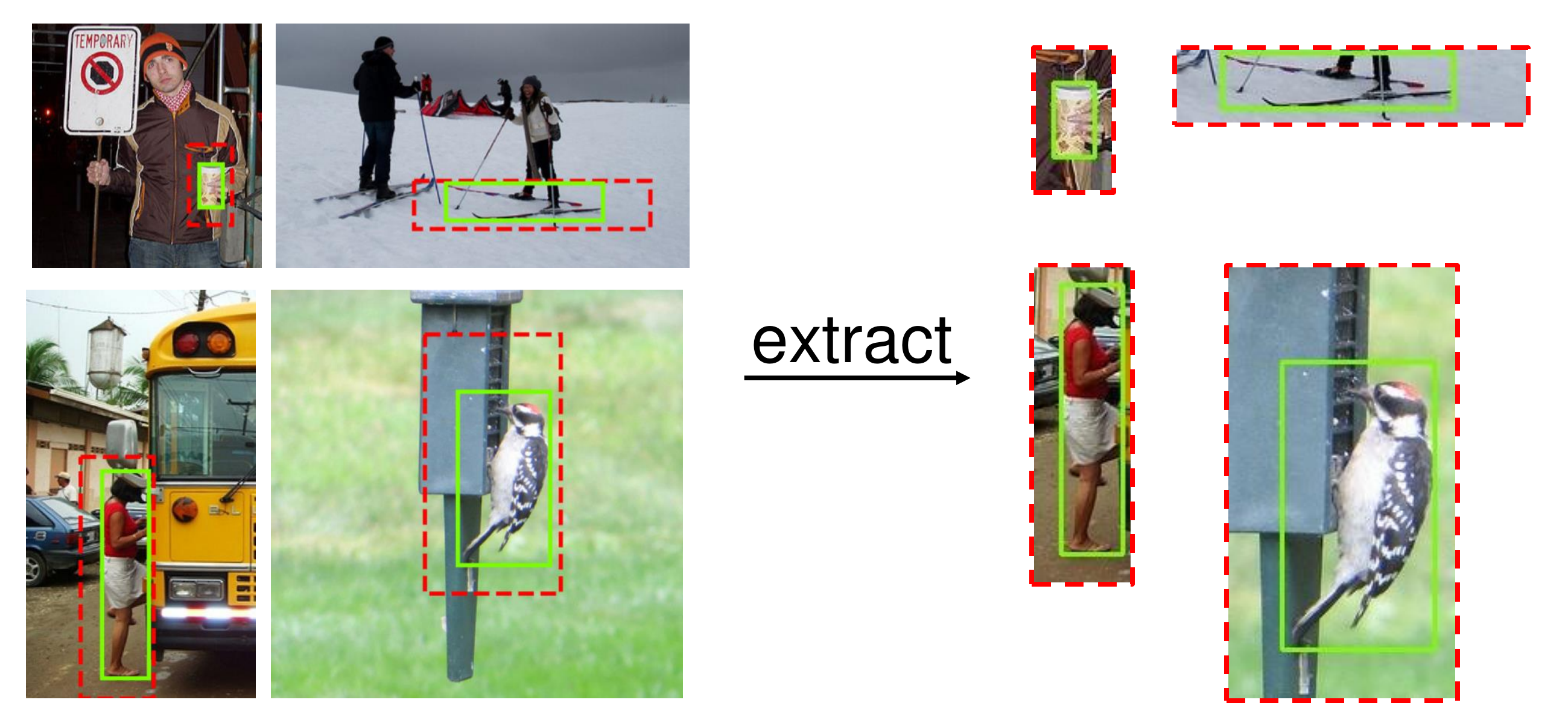}} \\
\subfigure[Warp]{
\label{warp_jigsaw}
\includegraphics[width=0.2\linewidth]{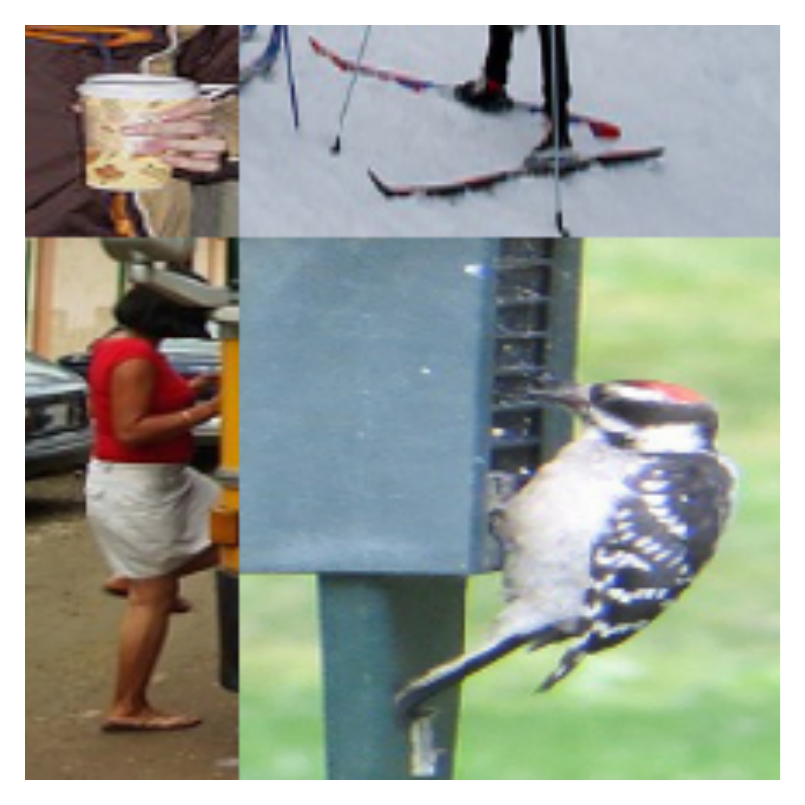}}
\subfigure[Resize]{
\label{resize_jigsaw}
\includegraphics[width=0.2\linewidth]{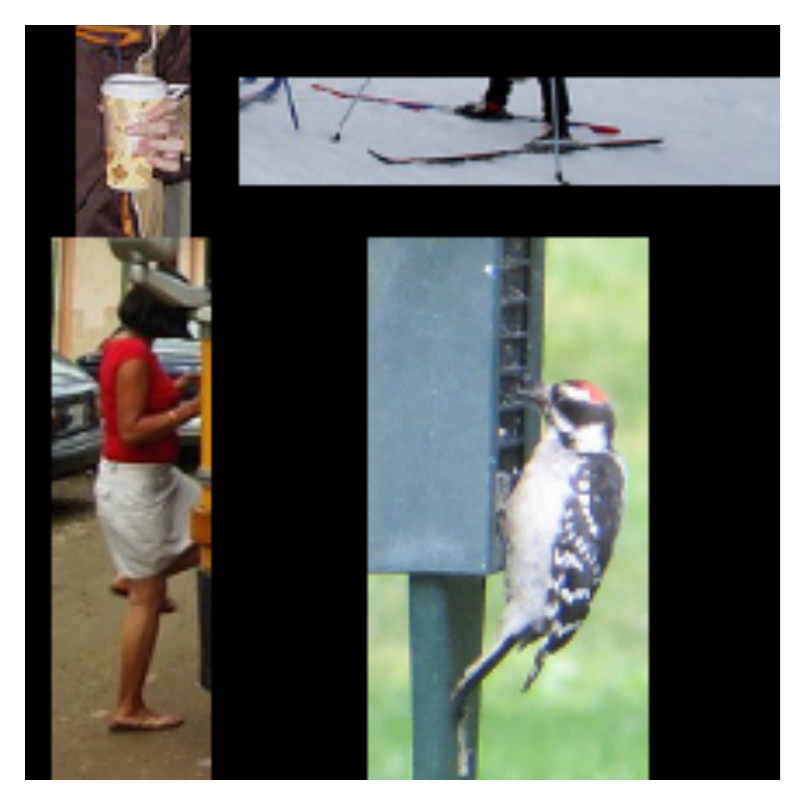}}
\subfigure[Crop]{
\label{crop_jigsaw}
\includegraphics[width=0.2\linewidth]{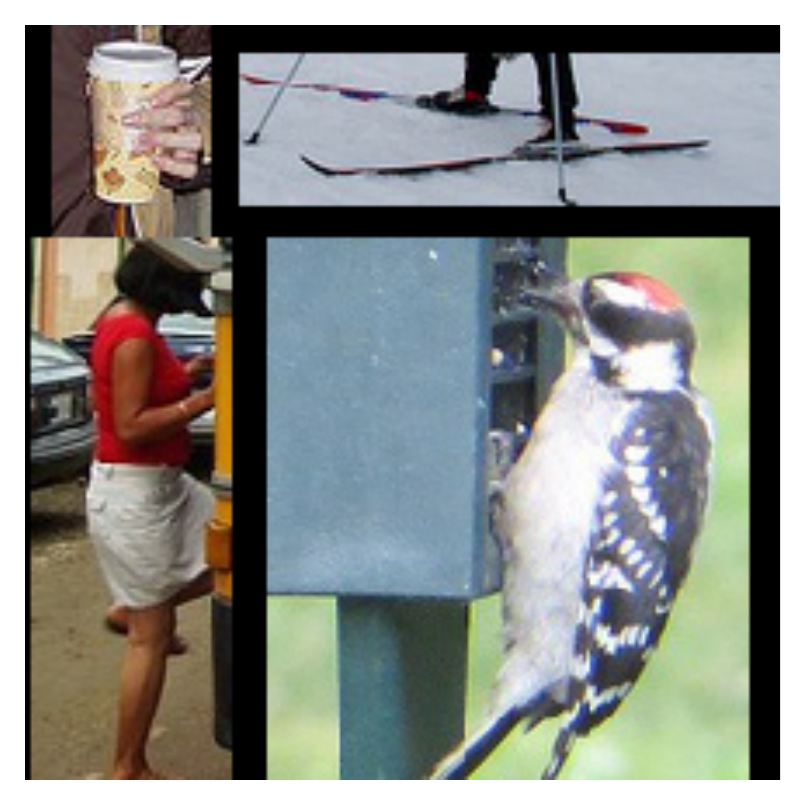}}
\caption{Visualization of three operations to adjust sample scale. (a) Samples are extracted from the original detection images. Similar to Fig.~2 in the main text, the boxes with green solid lines refer to Ground-Truth bounding boxes, those with red dash lines to the samples, which are randomly enlarged to incorporate more context information. (b) In `Warp', we change both the aspect ratio and scale of samples. (c) For `Resize', the aspect ratio is not changed and only the size is changed, then padding is applied to samples whose sizes are smaller than the pre-defined ones. (d) In `crop', we apply padding or random cropping to samples, conditioned on whether their sizes are smaller or larger than the pre-defined sizes}
\label{vis_adjust_scale}
\end{figure}

During the Montage assembled generation process, the samples are adjusted to fit the pre-defined size of in the template. The samples will be randomly cropped or zero-padded, which is conditioned on whether their sizes are smaller or larger than the pre-defined ones. Other possible solutions are to warp or resize the samples. Visualization examples of the different operations are depicted in Fig.~\ref{vis_adjust_scale}. From the examples, we can see that resize would result in too many pixels being uninformative and warping will distort the image. Crop is able to retain the shape and preserve more information, which makes it a better choice for size adjustment. Experiment results in Table~\ref{comp_resize_crop} also indicate that warp and resize lead to suboptimal performances compared with crop.

\setlength{\tabcolsep}{10pt}
\begin{table}
\begin{center}
\caption{Comparison of Warp, Resize and Pad \& Crop for scale adjustment during the pre-training process. The network structure is ResNet-50. The three pre-trained models are used for the subsequent detection training of Faster R-CNN~\cite{ren2015faster} and results are evaluated on COCO val2017. The results show that pad \& crop is more helpful for obtaining better pre-trained models. For the results of `Warp', we change both the size and aspect ratio while for those of `Resize', we only change the size}
\label{comp_resize_crop}
\begin{tabular}{llll}
\hline
Method & AP & AP$_{50}$ & AP$_{75}$ \\
\hline
Warp & 34.6 & 54.7 & 36.8 \\
Resize & 34.7 & 54.7 & 37.1 \\
Pad\&Crop & \textbf{35.2} & \textbf{55.7} & \textbf{37.6} \\
\hline
\end{tabular}
\end{center}
\end{table}
\setlength{\tabcolsep}{1.4pt}

\section{Visualization and comparison of different classification strategies}

\begin{figure}
\centering
\subfigure[Global classification]{
\label{global_loss}
\centering
\includegraphics[width=0.6\linewidth]{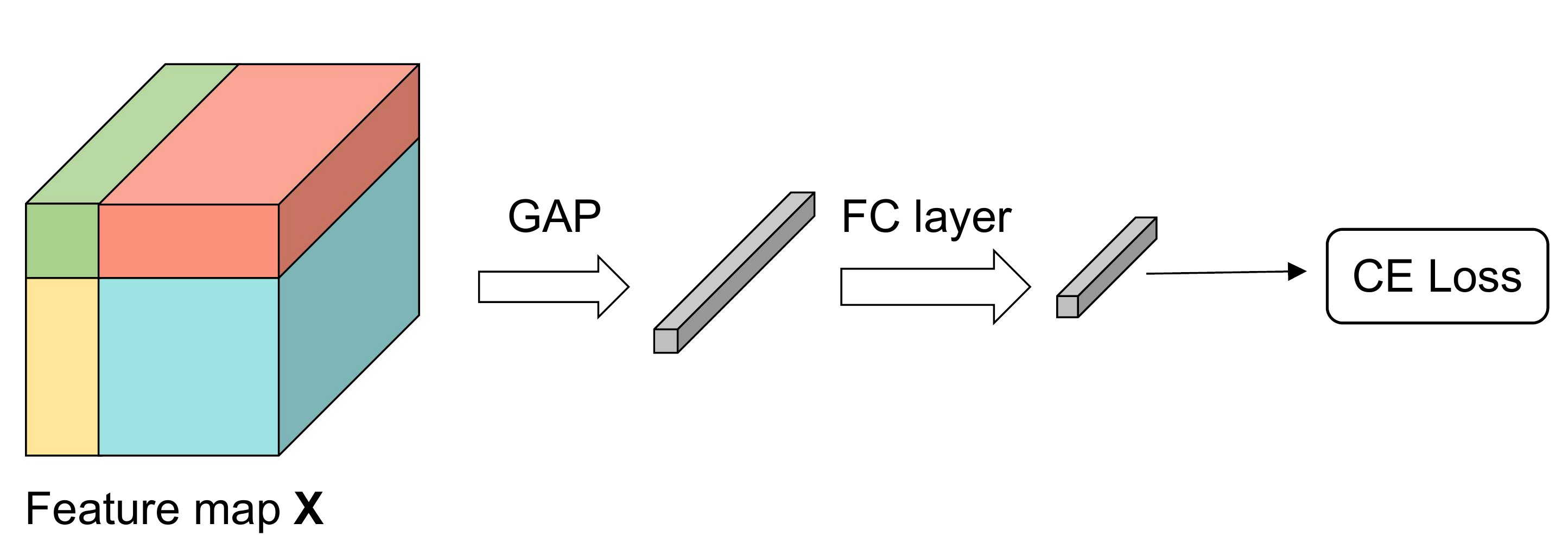}}
\subfigure[Block-wise classification]{
\label{part_loss}
\centering
\includegraphics[width=0.6\linewidth]{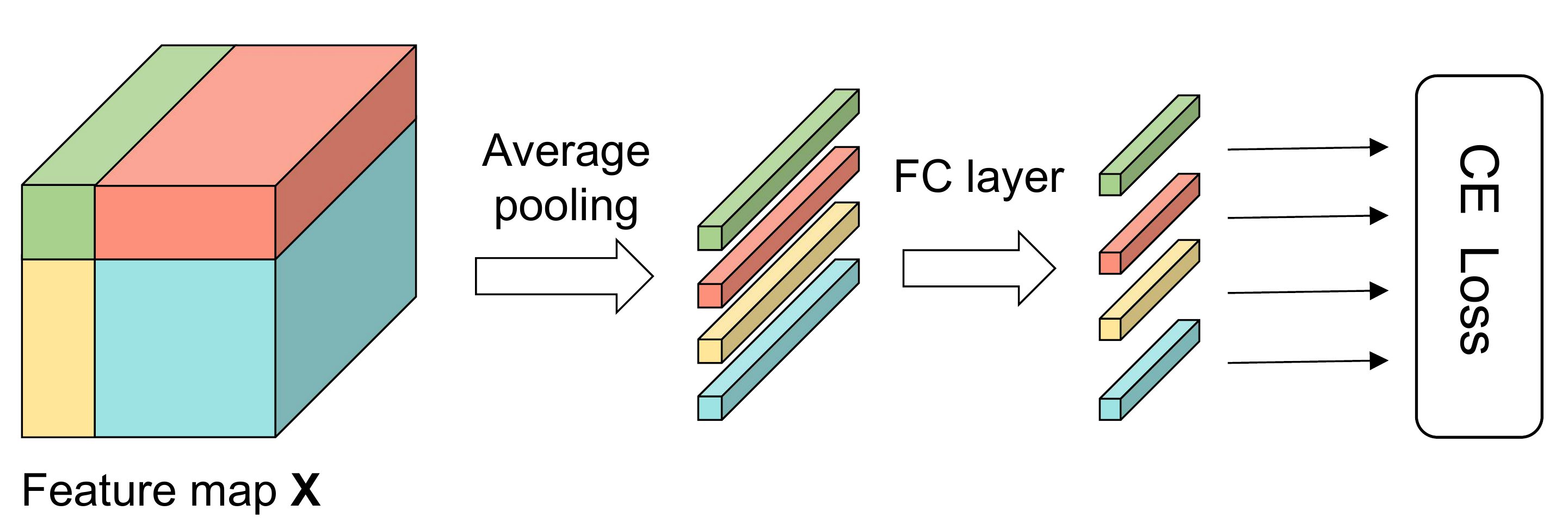}}
\caption{Visualization on the process of global classification (a) and block-wise classification (b). We use different colors to distinguish regions corresponding to the four samples. Best viewed in color}
\label{loss_process}
\end{figure}

This section provides visualization of global-wise and block-wise classification process, respectively, and also shows the comparison of different strategies.
The process of global classification is shown in Fig.~\ref{loss_process}(b), where global average pooling is exerted on the feature map and we assign the entire image a single global label. The global label is the weighted sum of labels of the four regions according to their region areas. Fig.~\ref{loss_process}(b) depicts the process of block-wise classification. Different from global classification, the average pooling is independently exerted on the four regions of feature map $\mathbf{X}$ corresponding to samples. Then we will apply classification operation on each region individually according to its label.

Table~\ref{comp_loss} present the experimental results on global-wise, block-wise and ERF-adaptive dense classification, respectively, from which we can see that the performance of the model pre-trained under our ERF-adaptive dense classification strategy is best among three strategies.

\setlength{\tabcolsep}{10pt}
\begin{table}
\begin{center}
\caption{Comparison of different classification strategies. The backbone CNN is ResNet50 and detection framework is Faster R-CNN~\cite{ren2015faster}. All results are evaluated on COCO val2017. The results show that the ERF-adaptive dense classification strategy clearly outperforms the other two strategies}
\label{comp_loss}
\begin{tabular}{llll}
\hline\noalign{\smallskip}
Strategy & AP & AP$_{50}$ & AP$_{75}$ \\
\noalign{\smallskip}
\hline
\noalign{\smallskip}
Global Cls. & 34.3 & 54.5 & 36.3 \\
Block-wise Cls. & 35.2 & 55.7 & 37.6 \\
ERF-adaptive Dense Cls. & \textbf{36.3} & \textbf{56.5} & \textbf{38.9} \\
\hline
\end{tabular}
\end{center}
\end{table}
\setlength{\tabcolsep}{1.4pt}

\section{Implementation details}
This section provides details of data augmentation during pre-training and training settings of detectors.

\noindent\textbf{Pre-training Augmentation.} The samples will first be resized according to a resize ratio chosen randomly from $\left[0.8, 1.5\right]$. Both the height and width will be adjusted by the same ratio so that its aspect ratio keeps unchanged. Random horizontal flip with probability 0.5 is also applied on each sample before being assembled into the new image. During Montage assembly, random cropping or zero padding is used to adjust the samples to the pre-defined sizes. After the stitching, the channels of assembled image are normalized with mean $[0.485, 0.456, 0.406]$ and std $[0.229, 0.224, 0.225]$.

\noindent\textbf{Training Details of Detectors.} For fair comparisons, we adopt the same training settings on detection for both ImageNet pre-trained models and Montage pre-trained models. We train our models on MS-COCO train2017 split. If not specified, all models are trained for 13 epochs on 8 Tesla V100 GPUs with total batch size 16. We use SGD as the optimizer with momentum 0.9 and weight decay 0.0001. The initial learning rate is 0.02 and decreases by factor 0.1 at epoch 9 and 12. The batch normalization layers are frozen during  training. The images are resized to $1333 \times 800$ and randomly flipped with probability 0.5 for augmentation. 